\documentclass{article}

\usepackage{arxiv}

\usepackage[utf8]{inputenc} 
\usepackage[T1]{fontenc}    
\usepackage{hyperref}       
\usepackage{url}            
\usepackage{booktabs}       
\usepackage{amsfonts}       
\usepackage{nicefrac}       
\usepackage{microtype}      
\usepackage{cleveref}       
\usepackage{lipsum}         
\usepackage{graphicx}
\usepackage{natbib}
\usepackage{doi}
\usepackage{todonotes}
\usepackage{wrapfig}

\title{Fox-1: Open Small Language Model for Cloud and Edge}

\date{}

\author{
\bf Zijian Hu\thanks{Equal contributions}$^{1}$,
~ \bf Jipeng Zhang\footnotemark[1]$^{1,2}$,
~ \bf Rui Pan\footnotemark[1]$^{3}$,
~ \bf Zhaozhuo Xu$^{1}$,
~ \bf Shanshan Han$^{1}$,\\
~ \bf Han Jin$^{1}$,
~ \bf Alay Dilipbhai Shah$^{1}$,
~ \bf Dimitris Stripelis$^{1}$,
~ \bf Yuhang Yao$^{1}$,\\
~ \bf Salman Avestimehr$^{1}$,
~ \bf Tong Zhang\footnotemark[2]$^{1,3}$,
~ \bf Chaoyang He\thanks{Corresponding author}$^{1}$\\
  $^{1}$TensorOpera Inc.
  \\ 
  $^{2}$The Hong Kong University of Science and Technology
  \\
  $^{3}$University of Illinois Urbana-Champaign\\
  \texttt{\{zjh,zhaozhuo,shanshan,rapahel,alay,dimitris,yuhang,avestimehr,ch\}@tensoropera.com}\quad \\
  \texttt{jzhanggr@ust.hk} \quad
  \texttt{ruip4@illinois.edu}\quad
  \texttt{tozhang@illinois.edu}
  \\
}

\renewcommand{\undertitle}{}

\begin{document}
\maketitle
\begin{abstract}
We present Fox-1, a series of small language models (SLMs) consisting of \href{https://huggingface.co/tensoropera/Fox-1-1.6B}{Fox-1-1.6B} and \href{https://huggingface.co/tensoropera/Fox-1-1.6B-Instruct-v0.1}{Fox-1-1.6B-Instruct-v0.1}. These models are pre-trained on 3 trillion tokens of web-scraped document data and fine-tuned with 5 billion tokens of instruction-following and multi-turn conversation data. Aiming to improve the pre-training efficiency, Fox-1-1.6B model introduces a novel 3-stage data curriculum across all the training data with 2K-8K sequence length. In architecture design, Fox-1 features a deeper layer structure, an expanded vocabulary, and utilizes Grouped Query Attention (GQA), offering a performant and efficient architecture compared to other SLMs.
Fox-1 achieves better or on-par performance in various benchmarks compared to StableLM-2-1.6B, Gemma-2B, Qwen1.5-1.8B, and OpenELM1.1B, with competitive inference speed and throughput. The model weights have been released under the Apache 2.0 license, where we aim to promote the democratization of LLMs and make them fully accessible to the whole open-source community.

\textbf{Base Model}: \url{https://huggingface.co/tensoropera/Fox-1-1.6B} \\
\textbf{Chat Model}: \url{https://huggingface.co/tensoropera/Fox-1-1.6B-Instruct-v0.1}
\end{abstract}



\section{Introduction}

\begin{figure}[h]
\begin{center}
\includegraphics[width=0.9\linewidth]{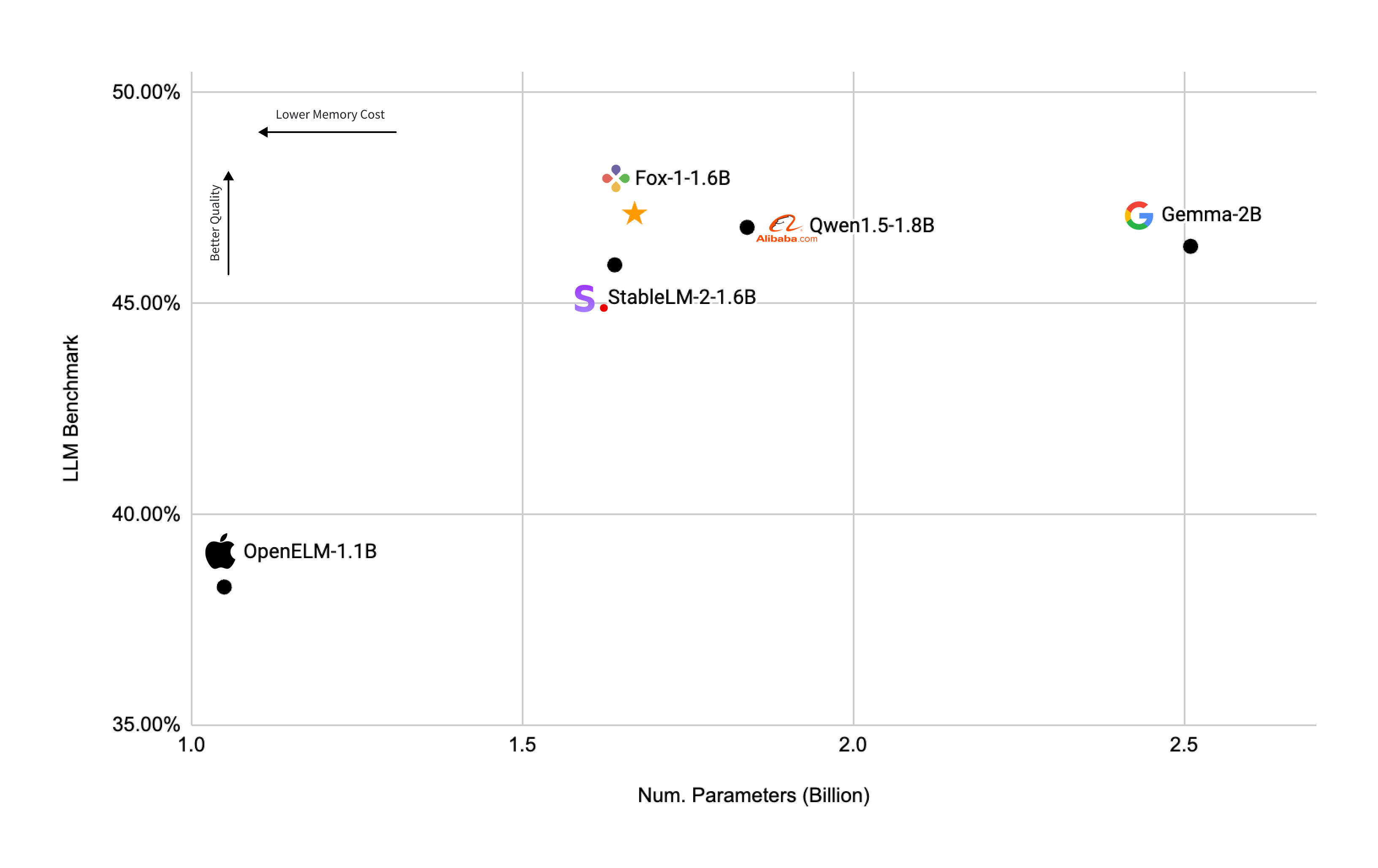}
\caption{Fox-1-1.6B compare with other SLMs.}
\label{fig:fox-overview}
\end{center}
\end{figure}

Recent advances in Large Language Models (LLMs) show that the unified next-token prediction paradigm excels in improving performance across diverse text generation tasks, such as code generation~\cite{chen2021humaneval}, math word problem solving~\cite{wei2022chainofthought_cot}, medical question answering~\cite{diao2023lmflow}. Furthermore, LLMs~\cite{brown2020gpt3gpt-3,openai2023gpt4} are widely acknowledged for their System 2 capabilities~\cite{deng2023RAR,saha2023branchsolvemerge,weston2023system2attention} in in-context learning~\cite{brown2020gpt3gpt-3} and chain-of-thought prompting~\cite{wei2022chainofthought_cot}. Generally, LLMs are developed by training a decoder-only transformer model on extensive text corpora~\cite{brown2020gpt3gpt-3,rae2021gopher}. Currently, most state-of-the-art models achieve high performance through fine-tuning white-box LLMs~\cite{touvron2023llama,touvron2023llama2,dubey2024llama3} or using black-box LLM APIs~\cite{openai2023gpt4,ouyang2022instructGPT}.

To further enhance the capabilities of decoder-only language models, LLM training focuses more on their scaling laws~\cite{kaplan2020scalinglaw} and compute-optimal training schedules. The key components include model size, the volume of training data, and the corresponding training compute. Several studies~\cite{hoffmann2022Chinchilla} provide guidance on optimally balancing the increase in model size with the amount of training data, consistently showing that large models require extensive datasets and substantial training compute.
Given that existing LLMs often reach hundreds of billions of parameters~\cite{touvron2023llama,touvron2023llama2,ouyang2022instructGPT,dubey2024llama3} for better performance, the training costs for these giant models are extremely high, leading to significant financial and environmental impacts. These costs also restrict most researchers from participating, and the high deployment costs further hinder the widespread application of these advanced AI technologies.

Recently, there has been a growing interest in training Small Language Models (SLMs) that achieve performance comparable to models four times their size. Examples of these models include the Phi series~\cite{DBLP:journals/corr/phi1,li2023phi1.5,abdin2024phi3}, TinyLlama~\cite{zhang2024tinyllama}, OpenELM~\cite{mehta2024openelm}, Gemma~\cite{team2024gemma}, MiniCPM~\cite{hu2024minicpm} and Qwen~\cite{bai2023qwen}. Research on these models explores various aspects of training effective SLMs, such as data paradigms, model architecture, and tokenizers. However, it remains unclear how to further enhance the training curriculum and data organization at each stage.

In this report, we introduce Fox-1, a series of Small Language Models (SLMs) that primarily explore research issues related to training curricula. We present a high-performance SLM with extensive investigation into training techniques. We release all our model weights publicly under the Apache 2.0 license, making them accessible on the TensorOpera AI Platform and Hugging Face \footnote{Base model is available at \url{https://huggingface.co/tensoropera/Fox-1-1.6B} and the instruction-tuned version is available at \url{https://huggingface.co/tensoropera/Fox-1-1.6B-Instruct-v0.1}}.

\section{Pre-Training}
\label{sec:pre_training}

Fox-1 pre-training involves curating and filtering a large corpus of open-sourced textual data, searching the model architecture, and developing the training recipe. In this section, we present the details of the above components.

\subsection{Pre-Training Data}

To pre-train Fox-1 we use 3 trillion tokens of textual data. To align with our proposed three-stage curriculum pre-training pipeline, we have reorganized the original data into three distinct collections. We have gathered an extensive range of textual data, encompassing unsupervised and instruction-tuning datasets, as well as diverse domains such as code, web content, mathematics, and science documents. To simplify the description of the data combination in the subsequent sections, we first list the dataset collection and then detail the data usage for each stage.

\paragraph{Raw Data Collection}

We collected datasets from publicly released large and high-quality sources, including Redpajama~\cite{together2023redpajama}, SlimPajama~\cite{cerebras2023slimpajama}, Dolma~\cite{soldaini2024dolma}, Pile~\cite{gao2020pile}, and Falcon~\cite{penedo2023falconrefinedweb} datasets. Here are the detailed datasets:
\begin{itemize}
    \item \textbf{Common Crawl.} We selected the Common Crawl (CC) split from all four datasets, which occupies almost 5T of storage. The SlimPajama-CC ranges from 2019 to 2023, Dolma-CC contains all the CC dumps before 2023, Pile-CC includes selected subsets from dumps across 2013 to 2020, and Falcon-CC ranges from 2008 to 2023. Deduplication was performed across all the CC subsets using a bloom filter algorithm~\cite{bloom1970bloomfilter}.
    \item \textbf{C4.} As demonstrated in the Llama-1 paper~\cite{touvron2023llama}, including diverse preprocessed Common Crawl datasets improves LLM performance. We also include the widely-recognized high-quality Common Crawl datasets C4~\cite{raffel2020C4_T5}.
    \item \textbf{Books.} Following the books collection in the Pile dataset, we include Books3~\cite{books3}, BookCorpus2~\cite{BookCorpus2}, and the Gutenberg~\cite{ProjectGutenberg19} collection in our Books dataset.
    \item \textbf{Wikipedia.} We selected Redpajama's Wiki collection, based on Wikipedia dumps from 2023-03-20, containing text in 20 different languages. This collection is similar to the Wiki collection in Llama.
    \item \textbf{Papers.} RedPajama, Dolma, and Pile all have their own paper collections. Dolma uses the pes2o~\cite{peS2o} dataset from AI2, which only contains abstracts of papers. For the paper collections in RedPajama and Pile, we also performed deduplication using a bloom filter algorithm.
    \item \textbf{Code.} Stack~\cite{kocetkov2022stack} is a 3.1T dataset containing permissively licensed source code across 30 languages. Since it maintains license permission checking, we include this as our code pre-training data collection.
    \item \textbf{Science Text.} We include PubMed, OpenWebText2~\cite{OpenWebtext2}, FreeLaw, USPTO Backgrounds, PhilPapers, and NIH Grant Abstracts from the Pile~\cite{gao2020pile} in our collection.
    \item \textbf{WebQ\&A.} The Reddit data in Dolma and StackExchange in SlimPajama are also collected separately to construct a web-based multi-round question answering dataset.
    \item \textbf{Math.} We include Open-Web-Math~\cite{paster2023openwebmath} and algerbraic-stack in the pre-training collection to enhance the model's mathematical abilities.
    \item \textbf{Instruction.} We collect several instruction or domain-specific datasets under permissive licenses to further increase the quality of the data.
    \item \textbf{RewriteBooks.} We select a subset of high-quality books and rewrite them with the Mixtral-8x22B~\cite{jiang2024mixtral} model to construct the RewriteBooks collection.
    \item \textbf{Others.} We got several other datasets widely used for training Language models like Flan~\cite{flan}, which is curated as an unification of various NLP taks.
\end{itemize}

\paragraph{First Stage Data}
This stage is comparable to the pre-training stage of existing large language models. For this stage, we randomly sampled half of the Common Crawl collection data to complete the first stage of pre-training. By initially using several Common Crawl-based datasets, we aim to create a more balanced distribution for the following training stages.

\begin{table}[h]
\centering
\begin{tabular}{lrrr}
\toprule
\textbf{Dataset}      & \textbf{Storage Size} & \textbf{Tokens} & \textbf{Average Seq Length} \\ \midrule
Common Crawl & 7.9TB          & 1.05T   & 591.29              \\ \bottomrule
\end{tabular}
\vspace{0.1in}
\caption{Dataset statistics for the first stage training of collection. Here, ``TB''  refers to terabytes for storage statistics, while ``T'' denotes the number of tokens in the respective training datasets.}
\label{table:stage1 data}
\end{table}

\paragraph{Second Stage Data}
In this stage, we use a smaller scale of training data while enhancing long-context capabilities. To achieve this, we collect data from more diverse domains. This includes another part of the Common Crawl collection, Books, C4, Math, Papers, Wiki, Code, and all subsets from Science Text and WebQ\&A.

\begin{table}[h]
\centering
\begin{tabular}{lrrrc}
\toprule
\textbf{Dataset} & \textbf{Storage Size} & \textbf{Num. Tokens} & \textbf{Seq. Length Avg.} & \textbf{Chunk Length} \\ \midrule
Common Crawl & 9.93 TB & 1.47 T & 1269.09 & 4K \\ 
The Stack & 1.89 TB & 375.98 B & 1783.49 & 4K \\ 
C4 & 853.92 GB & 169.75 B & 471.69 & 4K \\ 
Papers & 785.42 GB & 121.18 B & 2837.34 & 4K \\ 
Dolma\_Reddit & 424.61 GB & 79.32 B & 210.13 & 4K \\ 
PubMed\_Central & 227.16 GB & 45.35 B & 7984.65 & 8K \\
OpenWebText2 & 142.76 GB & 28.12 B & 869.78 & 4K \\ 
Books & 124.54 GB & 24.91 B & 121057 & 8K \\
RedPajama StackExchange & 102.49 GB & 20.15 B & 680.03 & 4K \\ 
Wikipedia & 101.49 GB & 20.03 B & 671.32 & 4K \\ 
Freelaw & 98.42 GB & 19.63 B & 3872.02 & 8K \\ 
Open-Web-Math & 66.59 GB & 13.22 B & 2093.74 & 4K \\
USPTO\_Backgrounds & 49.19 GB & 9.67 B & 869.23 & 4K \\ 
PhilPapers & 5.64 GB & 1.13 B & 17639.44 & 8K \\ 
NIH ExPorter & 3.72 GB & 720.67 M & 405.34 & 4K \\ 
\midrule
Total & 14.35 TB & 2.30 T & 1105.56 & - \\ \bottomrule
\end{tabular}
\vspace{0.1in}
\caption{Dataset statistics for the second stage training of collection. Here, ``TB'' and ``GB'' indicate terabytes and gigabytes for storage space statistics, while ``T'', ``B'', ``M'', and ``K'' denote the number of tokens in the respective training datasets.}
\label{table:stage2 data}
\end{table}

\paragraph{Third Stage Data}
In the final stage, we aim to curate an extremely high-quality data collection. As demonstrated in Phi-family models~\citep{gunasekar2023phi1,li2023phi1.5}, machine-generated datasets are incorporated at this stage. Consequently, we include the Instruction and RewriteBooks subsets in this stage's data collection. We also down-sample a subset with 100B tokens from the second stage data mix for this stage.


\begin{table}[h]
\centering
\begin{tabular}{lrrr}
\toprule

\textbf{Dataset} & \textbf{Final Size} & \textbf{Num. Tokens} & \textbf{Ratio} \\ 
\midrule
Instruction & 183.87 GB & 36.02 B & 34.07\% \\ 
The Stack & 94.34 GB & 18.79 B & 17.78\% \\ 
flan  & 29.05 GB & 15.47 B & 14.63\% \\ 
algebraic-stack & 10.92 GB & 12.51 B & 11.83\% \\ 
Papers & 30.63 GB & 6.06 B & 5.73\% \\ 
Books & 7.75 GB & 5.04 B & 4.77\% \\
Dolma\_Reddit & 20.14 GB & 3.96 B & 3.74\% \\
PubMed Central & 11.66 GB & 2.26 B & 2.14\% \\ 
OpenWebText2 & 7.35 GB & 1.40 B & 1.33\% \\
RedPajama\_StackExchange & 5.36 GB & 1.00 B & 0.95\% \\ 
Wikipedia & 5.33 GB & 997.75 M & 0.94\% \\ 
Freelaw & 5.23 GB & 977.90 M & 0.93\% \\ 
Open-Web-Math & 3.63 GB & 657.70 M & 0.62\% \\ 
USPTO\_Backgrounds & 2.74 GB & 479.97 M & 0.45\% \\
PhilPapers & 601.02 MB & 52.96 M & 0.05\% \\
NIH ExPorter & 499.56 MB & 32.65 M & 0.03\%
\\ \midrule
\textbf{Total} & \textbf{419.07 GB} & \textbf{105.71 B} & \textbf{100.00\%} \\ \bottomrule
\end{tabular}
\vspace{0.1in}
\caption{Dataset statistics for the second stage training of collection. Here, ``TB'', ``GB'' and ``MB'' refer to terabytes, gigabytes and megabytes for storage space statistics, while ``B'' and ``M'' denote the number of tokens in the respective training datasets.}
\label{table:stage3 data}
\end{table}

\subsubsection{Tokenization}

We hypothesize that it is easier to correctly predict a few consecutive tokens than many consecutive tokens~\cite{le2023bloom,team2024gemma,tao2024scalingvocab,kudo2018sentencepiece}.
Therefore, we believe using a large vocabulary could lead to better downstream performance.
Arguably, a large vocabulary typically yields fewer tokens for a given text corpus~\cite{le2023bloom,team2024gemma}, which could lead to better inference performance.

Although Fox-1 is an English-only SLM,  we aim to select a sufficiently general vocabulary to leverage intermediate checkpoints for future multilingual capabilities and domain-specific applications.
We selected the tokenizer from \citet{team2024gemma}, which offers a vocabulary size of 256K, making it one of the largest available at the time of this report.

Increasing the size of vocabulary has at least two major benefits. First, the effective context length is implicitly extended given the denser information encoded in each token. For example, a vocabulary with size $26$ can only encode one character in [a-z], but a vocabulary with size $26^2$ can encode two alphabetical letters at a time, leading to longer representable strings with a fixed length of tokens. Second, a larger vocabulary size reduces the probability of unknown words or phrases, resulting in better downstream task performance in practice.

\subsubsection{Model Architecture}

Fox-1 employs a decoder-only transformer architecture inspired by \cite{touvron2023llama2,touvron2023llama,dubey2024llama3,liu2024mobilellm} with 1.6B total parameters while introducing various improvements and redesigns for better performance.

\paragraph{Deeper Network}
A fundamental trade-off in model architecture design is depth versus width. While wider and shallower networks allow better memorization, deeper and thinner networks present stronger reasoning ability~\citep{cheng2016wide,liu2024mobilellm}. In accordance with this principle, Fox-1 uses a deeper architecture than most modern SLMs. Specifically, Fox-1 consists of 32 self-attention layers, which is 78\% deeper than Gemma-2B (18 layers), 33\% deeper than StableLM-2-1.6B (24 layers), and 33\% deeper than Qwen1.5-1.8B (24 layers).

\paragraph{Shared Embedding} Fox-1 utilizes a large vocabulary of 256,000 with a hidden dimension of 2,048, resulting in approximately 0.5 billion parameters. Larger models typically use separate embedding layers for the input (vocabulary size to embedding size) and output (embedding size to vocabulary size) layers.
For Fox-1, the embedding layers alone would require 1 billion parameters. To reduce the total number of parameters, we follow the approach of \citet{zhang2022opt,liu2024mobilellm} and share the input and output embedding layer, maximizing weight utilization and reducing the parameter count by 0.5 billion, approximately 30.49\% of the final model size.

\paragraph{Pre-normalization} Similar to~\citet{touvron2023llama2}, we use RMSNorm~\citep{zhang2019rootmeansquarelayer} to normalize the input of each transformer layer. RMSNorm is the dominant choice of pre-normalization in modern large language models~\citep{team2024gemma,bai2023qwen,touvron2023llama2}, where it demonstrates better efficiency than LayerNorm~\citep{ba2016layernorm}.

\begin{table}[h]
\centering
\begin{tabular}{ll}
\toprule
\textbf{Feature} & \textbf{Fox-1} \\ 
\midrule
Parameters & 1.6B \\
Attention Mechanism & GQA~\citep{ainslie2023gqa} \\
Non-linearity & SwiGLU~\citep{shazeer2020gluvariantsimprovetransformer} \\
Normalization & RMSNorm~\citep{touvron2023llama2} \\
Vocabulary Size & 256K \\
Context length & 8K \\
Embedding Share & True \\
\bottomrule
\end{tabular}
\vspace{0.1in}
\caption{Fox-1 Model Architecture.}
\label{table:model_architecture}
\end{table}


\paragraph{Rotary Positional Embeddings (RoPE)} Fox-1 accepts at most 8K-length input tokens by default. To improve the performance for a longer context window, we employ the widely adopted Rotary Position Embedding (RoPE)~\citep{su2024rope} with $\theta$ set to $10,000$ to facilitate the encoding of relative positional dependency between tokens.

\paragraph{Grouped Query Attention (GQA)}
Grouped Query Attention (GQA)~\citep{ainslie2023gqa} divides the query heads of multi-head attention layers into groups where each group shares the same set of key-value heads.
We use GQA~\citep{ainslie2023gqa} with 4 key-value heads and 16 attention heads to improve training and inference speed and reduce memory usage. 

\subsubsection{Training}

The pre-training of long sequences is known to be challenging given the training inefficiency incurred by the quadratic complexity of the attention mechanism~\citep{vaswani2017transformers}. To mitigate this problem, a 3-stage curriculum learning strategy is introduced in the pre-training stage of Fox, where the chunk length of the training sample is gradually increased from 2K to 8K to ensure the long-context ability at a small cost.

\begin{table}[h]
\centering
\begin{tabular}{lrrr}
\toprule
\textbf{Stages} & \textbf{Dataset Size} & \textbf{Chunk length} & \textbf{Batch Size} \\ 
\midrule
Stage 1 & 1.05T & 2K & 2M Tokens \\
Stage 2 & 1.58T & 4K-8K & 4M Tokens \\
Stage 3 & 0.06T & 4K-8K & 4M Tokens \\
\bottomrule
\end{tabular}
\vspace{0.1in}
\caption{Fox-1 Pre-training Sequence Length Curriculum: In this context, ``T'' represents the total number of tokens in the respective training datasets. Chunk length specifies the number of tokens per sample within a mini-batch. It is worth noting that the actual number of tokens used slightly differs from previous counts in the dataset section. This discrepancy arises because a subset was sampled with a limited number of training steps.}
\label{table:data_curriculum}
\end{table}

Specifically, stage 1 comprises $\sim 39 \%$ total data samples in the whole pre-training process, where the 1.05T-token dataset is chunked into 2K-length samples, with a batch size 2M tokens. We use 2,000 steps of linear warm-up for this stage.
Stage 2 includes $\sim 59 \%$ samples with 1.58T tokens and increases the chunk length from 2K to 4K and 8K. The actual chunking lengths vary across different data sources, which are decided based on the natural average lengths in each data source. The batch size also grows to 4M for improving the training efficiency, given stage 2 being the most time-consuming stage with diverse sources of different datasets.
Finally, in stage 3, Fox is trained with 62B tokens ($\sim 0.02\%$) of high-quality data to lay the groundwork for different downstream task abilities, such as instruction-following, chitchat, domain-specific question-answer, etc.
Across all stages, we utilize AdamW and WSD schedule~\citep{hu2024minicpm} to train Fox, along with learning rate $4 \times 10^{-4}$, weight decay $0.1$, AdamW $\beta_1 = 0.9, \beta_2 = 0.95$, WSD temperature $16,000$.










\section{Experimental Results}

Following the evaluation setup of the Open LLM Leaderboard~\citep{open-llm-leaderboard}, we evaluated Fox-1 and other SLMs on ARC Challenge (25-shot)~\citep{clark2018think}, HellaSwag (10-shot)~\citep{zellers2019hellaswag}, TruthfulQA (0-shot)~\citep{lin2021truthfulqa}, MMLU (5-shot)~\citep{hendrycks2020measuring}, Winogrande (5-shot)~\citep{sakaguchi2021winogrande}, and GSM8k (5-shot)~\citep{cobbe2021gsm8k} on the a machine with 8$\times$H100 GPUs and reported the average score of the 6 benchmarks.

\begin{figure}[h!]
\begin{center}
\includegraphics[width=0.9\linewidth]{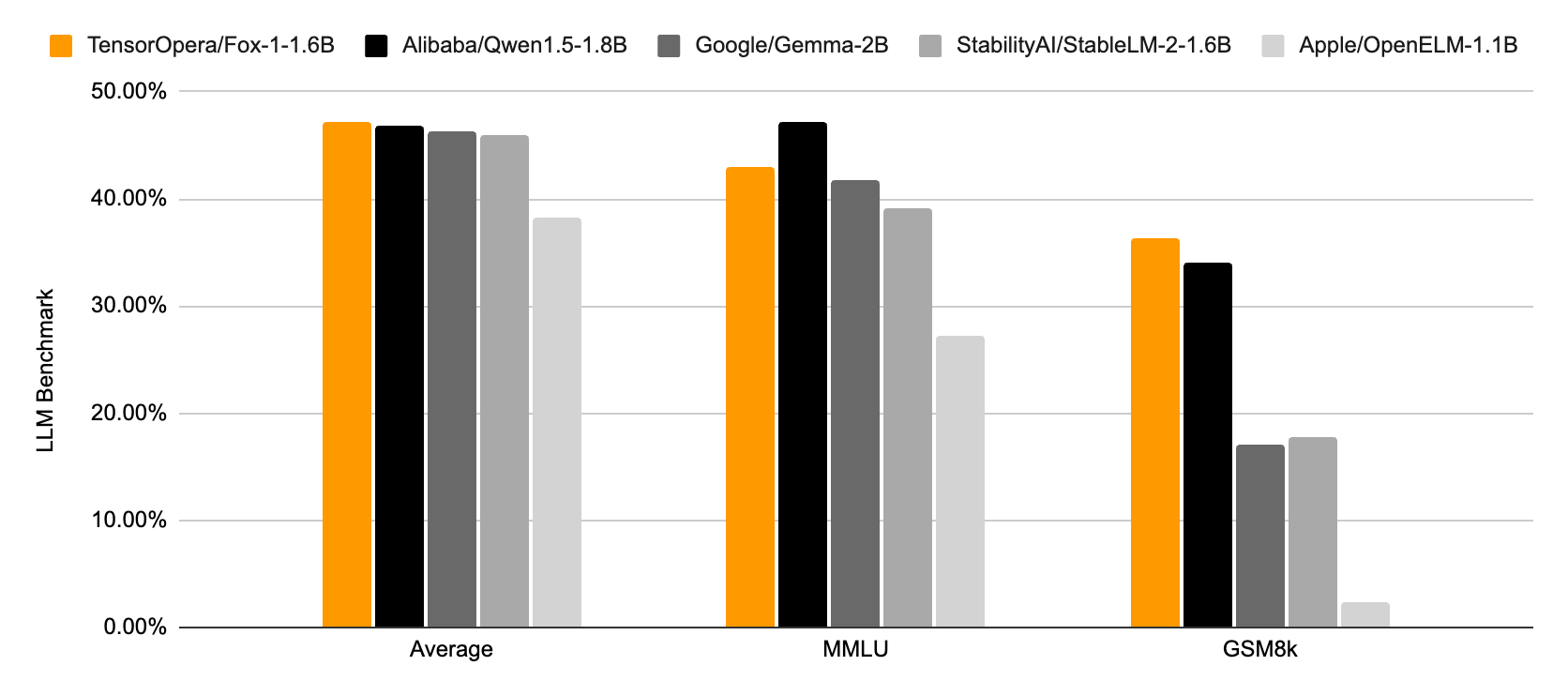}
\caption{Performance compared to other SLMs.}
\label{fig:fox-benchmarks}
\end{center}
\end{figure}

\begin{table}[h!]
\centering
\begin{tabular}{lcccc}
\toprule
\textbf{Model} & \textbf{Parameters} & \textbf{Average score*} & \textbf{Math (GSM8k)} & \textbf{World Knowledge (MMLU)} \\ \midrule
\textbf{TensorOpera/Fox-1-1.6B} & 1.67B & \textbf{47.13\%} & \textbf{36.39\%} & 43.05\% \\
Alibaba/Qwen-1.5-1.8B & 1.84B & 46.81\% & 34.04\% & \textbf{47.15\%} \\
Google/Gemma-2B & 2.51B & 46.36\% & 17.06\% & 41.15\% \\
StabilityAI/StableLM-2-1.6B & 1.64B & 45.92\% & 17.74\% & 39.16\% \\
Apple/OpenELM-1.1B & 1.05B & 38.28\% & 2.27\% & 27.28\% \\ \bottomrule
\end{tabular}
\vspace{0.1in}
\caption{Fox-1 performance compared to other SLMs. *Average score: the average of ARC, HellaSwag, MMLU, GSM8k, TruthfulQA, and Winograde.}
\label{table:fox-benchmarks}
\end{table}

As shown in figure \ref{fig:fox-benchmarks} and table \ref{table:fox-benchmarks}, Fox-1 performs better than or on par with Gemma-2B~\citep{team2024gemma}, Qwen1.5-1.8B~\citep{bai2023qwen}, StableLM-2-1.6B~\citep{bellagente2024stable}, and OpenELM1.1B~\citep{mehta2024openelm} across standard LLM benchmarks. For GSM8k, Fox-1 achieves 36.39\%, outperforming all baselines. Fox-1 also surpasses  Gemma-2B, StableLM-2-1.6B, and OpenELM 1.1B on MMLU despite being only half the size of Gemma-2B.

\begin{figure}[h!]
\begin{center}
\includegraphics[width=0.9\linewidth]{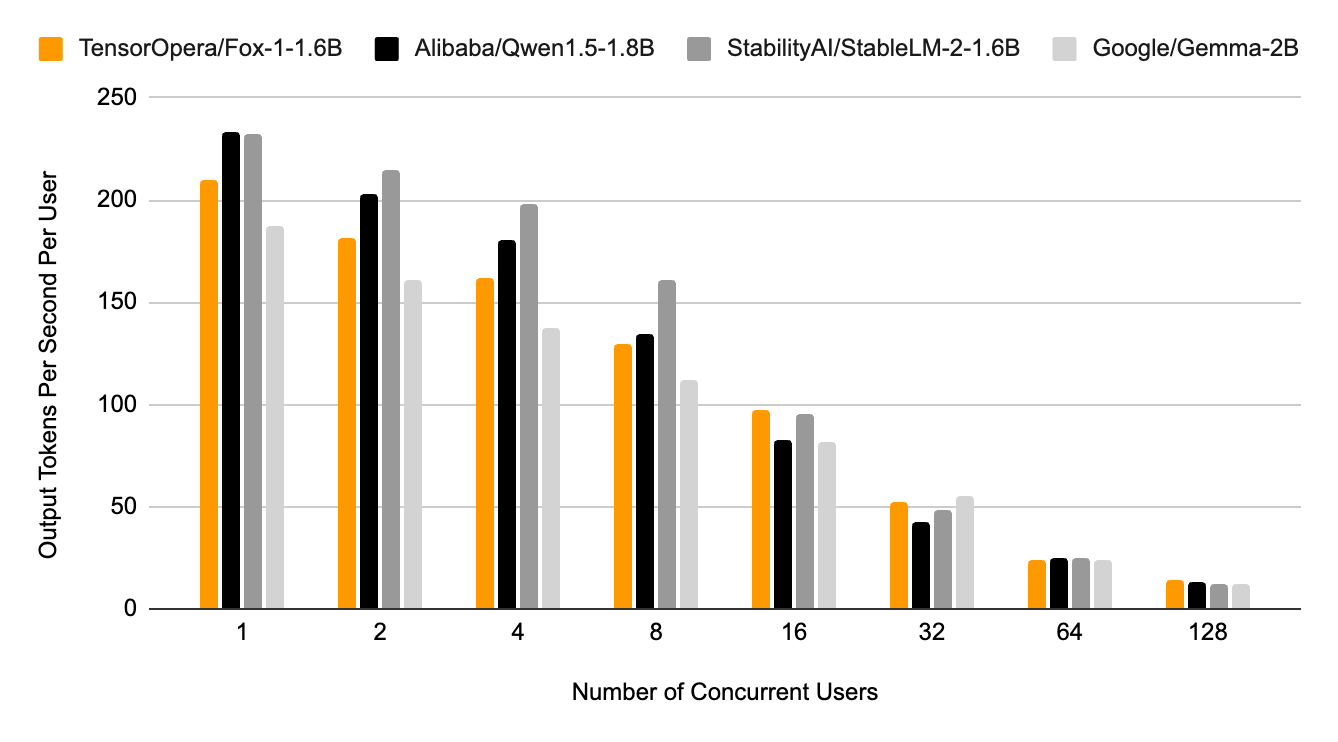}
\caption{Inference speed compared to other SLMs.}
\label{fig:fox-inference-speed}
\end{center}
\end{figure}

\paragraph{Inference Efficiency of Fox-1}
We evaluated the end-to-end inference efficiency of Fox-1, Qwen1.5-1.8B, and Gemma-2B using vLLM with the TensorOpera serving platform on a single NVIDIA H100 in BF16 precision.
To simulate real-world usage in multi-user scenarios, we use the OpenOrca~\citep{OpenOrca} dataset and send concurrent requests to the same inference server.
Performance was measured as output tokens per user per second, with each request averaging 234 tokens and each response 512 tokens.
As shown in figure \ref{fig:fox-inference-speed}, Fox-1 achieves a throughput exceeding 200 tokens per second, surpassing Gemma-2B and matching Qwen1.5-1.8B in the same deployment environments.
This high throughput can be attributed to Fox-1's architectural design, which incorporates Grouped Query Attention (GQA) for efficient query processing.

We did not include OpenELM, as it is unsupported by vLLM. With BF16 precision, Fox-1 only needs 3703MiB of GPU Memory, while Qwen1.5-1.8B, StableLM-2-1.6B, and Gemma-2B, respectively, requires 4739MiB, 3852MiB, and 5379MiB.

\section{Conclusion}
The Fox-1 series of small language models, including Fox-1-1.6B and Fox-1-1.6B-Instruct-v0.1, show advancement in efficient pre-training and instruction-following capabilities. Through the 3-stage data curriculum and a deep architecture with a large vocabulary size, Fox-1 excels across various benchmarks, outperforming or matching other small language models like StableLM-2-1.6B and Gemma-2B. The success of Fox-1 demonstrates the possibility of pre-training language models with competitive performance even with limited data resources.

\bibliographystyle{unsrtnat}
\bibliography{references}  






\end{document}